\documentclass[runningheads]{llncs}

\usepackage{eccv}

\usepackage{eccvabbrv}
\usepackage{adjustbox}
\usepackage{array}
\usepackage{ragged2e}
\usepackage[ruled,vlined]{algorithm2e}
\newcolumntype{C}[1]{>{\centering\arraybackslash}p{#1}}

\usepackage[table]{xcolor} 
\newcommand{\red}[1]{\textcolor{red}{#1}}
\usepackage{colortbl}      
\usepackage{amssymb}
\usepackage{graphicx}
\usepackage{booktabs}
\usepackage{amsmath}  
\usepackage{amsfonts} 
\usepackage{bm}       
\usepackage{wrapfig}
\usepackage{algpseudocode}
\usepackage[accsupp]{axessibility}  

\usepackage{hyperref}
\usepackage{multirow}

\usepackage{orcidlink}

\newcommand{\Yes}{\checkmark}
\newcommand{\No}{$\times$}

\definecolor{C1}{RGB}{30,70,110}
\definecolor{C2}{RGB}{31,120,180}
\definecolor{C3}{RGB}{106,172,217}

\newcommand{\Tref}[1]{Table~\ref{#1}}
\newcommand{\Fref}[1]{Figure~\ref{#1}}
\newcommand{\Eref}[1]{Equation~\ref{#1}}

\definecolor{richpurple}{RGB}{123, 63, 180}
\definecolor{mauve}{RGB}{186, 94, 134}
\definecolor{softcoral}{RGB}{250, 128, 114}

\begin{document}

\title{HiPolicy: Hierarchical Multi-Frequency Action Chunking for Policy Learning} 

\titlerunning{Abbreviated paper title}


\author{Jiyao Zhang\inst{1, 2} \and
Zimu Han\inst{3} \and
Junhan Wang\inst{1} \and
Xionghao Wu\inst{4} \and
Shihong Lin\inst{5} \and
Jinzhou Li\inst{1} \and
Hongwei Fan\inst{1, 2} \and
Ruihai Wu\inst{1} \and
Dongjiang Li\inst{6} \and
Hao Dong\inst{1, 2}}

\authorrunning{J. Zhang, et al.}
\titlerunning{HiPolicy}

\institute{
\scriptsize
\mbox{CFCS, School of CS, PKU} \and
National Key Laboratory for Multimedia Information Processing, School of CS, PKU \and 
\mbox{XJTU \quad \and THU \quad \and BUAA \quad \and Jingdong Technology Information Technology Co., Ltd} \\
\email{jiyaozhang@stu.pku.edu.cn} \\
\vspace{5pt}
\small
\url{https://hipolicy.github.io}
}

\maketitle

\begin{abstract}
  Robotic imitation learning faces a fundamental trade-off between modeling long-horizon dependencies and enabling fine-grained closed-loop control. Existing fixed-frequency action chunking approaches struggle to achieve both. Building on this insight, we propose \textbf{HiPolicy}, a hierarchical multi-frequency action chunking framework that jointly predicts action sequences at different frequencies to capture both coarse high-level plans and precise reactive motions. We extract and fuse hierarchical features from history observations aligned to each frequency for multi-frequency chunk generation, and introduce an entropy-guided execution mechanism that adaptively balances long-horizon planning with fine-grained control based on action uncertainty. Experiments on diverse simulated benchmarks and real-world manipulation tasks show that HiPolicy can be seamlessly integrated into existing 2D and 3D generative policies, delivering consistent improvements in performance while significantly enhancing execution efficiency.
\end{abstract}
\section{Introduction}
\label{sec:intro}

\begin{wrapfigure}{r}{0.5\textwidth}
    \centering
    \vspace{-36pt}
    \includegraphics[width=0.5\textwidth]{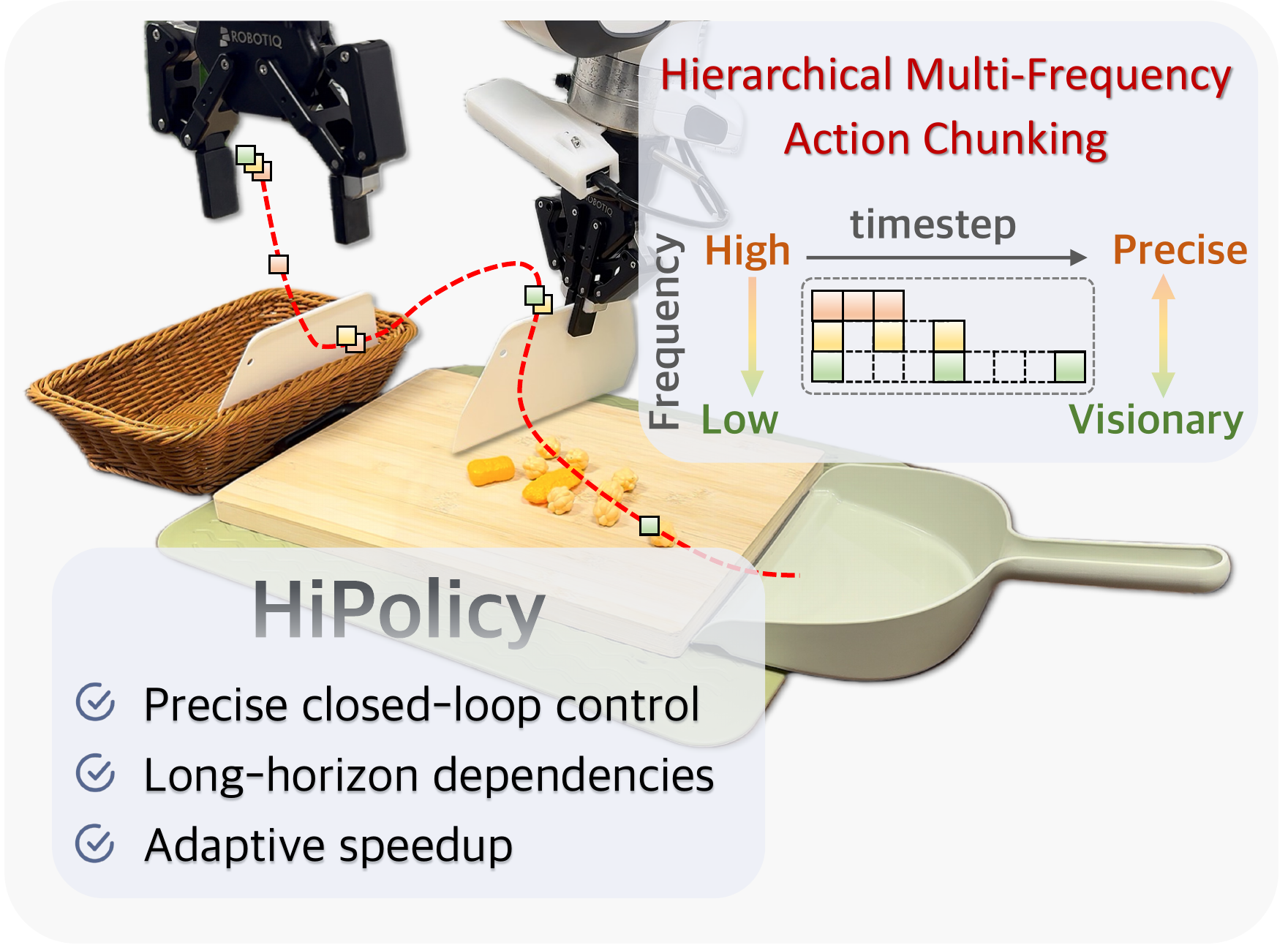} 
    \vspace{-16pt}
    \caption{We propose \textbf{HiPolicy}, a hierarchical multi-frequency action chunking for policy learning, modeling long-horizon dependency and precise closed-loop control.
    }  
    \vspace{-30pt}
    \label{fig:teaser}
\end{wrapfigure}

\begin{figure}[t]    
    \centering
    \includegraphics[width=1.0\textwidth]{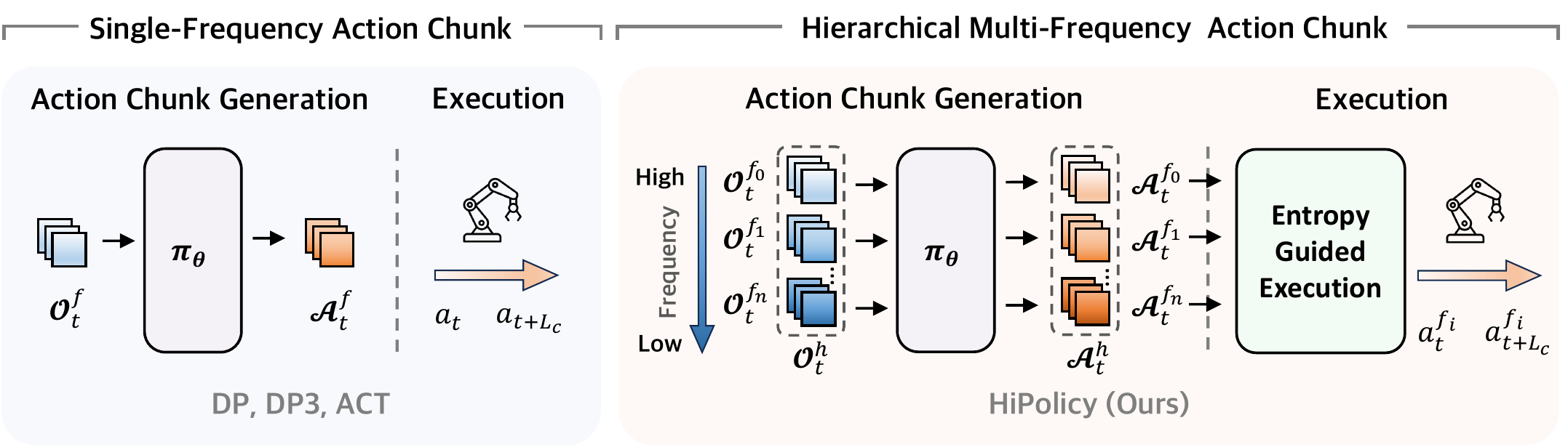}
    \captionof{figure}{
        \textbf{Comparison of HiPolicy with Existing Methods.} Existing imitation learning methods typically predict an action chunk at a single fixed frequency, leading to a trade-off between long-horizon dependency modeling and precise closed-loop control. In contrast, HiPolicy proposes hierarchical multi-frequency action chunking, enabling the capture of both long-term intentions and precise closed-loop adjustments. Additionally, the proposed action-entropy guided adaptive execution mechanism selects action frequency based on action uncertainty, balancing robustness and efficiency during task execution.
    }
    \label{fig:different_methods}
\end{figure}

Learning from human demonstrations has emerged as a powerful paradigm for robotic manipulation \cite{Wang2024RISE3P, Chi2023DiffusionPV, Ze2024DP3, Ke20243DDA, xue2025reactive, Zhang2025ChainofActionTA, Gong2024CARPVP, Su2025DensePB}, enabling policies to acquire complex skills without explicit reward engineering or exhaustive exploration. Imitation Learning (IL) \cite{andrychowicz2018learning, finn2017learning} has shown promise in both task-specific policy training and general-purpose vision-language-action (VLA) model \cite{Intelligence202505AV, internvlam1}, development, providing a scalable route toward versatile robot control \cite{Chi2023DiffusionPV, kim24openvla, tian2024predictive, Black20240AV}. Among various IL approaches, generative policies such as Diffusion Policy \cite{Chi2023DiffusionPV}, ACT \cite{Zhao2023LearningFB}, and DP3 \cite{Ze2024DP3} have become dominant due to their ability to model continuous action distributions with high temporal fidelity, delivering strong performance across diverse manipulation tasks.  

Despite the effectiveness of imitation learning, it still faces fundamental challenges. First, error accumulation remains a critical issue. Small discrepancies in predicted actions can compound over time, leading to substantial state deviations and eventual task failure \cite{kelly2019hg, ross2011reduction, laskey2017dart}. Second, modeling long-horizon dependencies is hard, especially for non-Markovian behaviors frequently observed in human demonstrations, such as pauses or subtle oscillatory motions \cite{Li2025CogVLACV}. Recent work has sought to address this via action chunking \cite{Zhao2023LearningFB}, predicting multi-step sequences instead of single-step actions. While this improves temporal coherence, its effectiveness is tightly coupled to two hyperparameters: action frequency and chunk size. As shown in~\Fref{fig:teaser}, at the same chunk size, low-frequency chunks capture long-horizon dependencies but lack the temporal resolution for fine-grained closed-loop control, whereas high-frequency chunks offer fine-grained adjustments but are less effective at modeling long-horizon dependencies.  This trade-off presents a significant challenge in designing effective action chunking strategies for robotic policies.

Human motor control studies reveal that complex movements naturally integrate components operating at multiple frequencies \cite{Flanagan2006ControlSI,Taniguchi2025System0Q, miller2024timescales}. Long-duration, low-frequency motions encode high-level goals and stage intentions, while short-duration, high-frequency motions enable precise adjustments and reactive control. 
This raises a key question: 
\textit{Could such a hierarchical multi-frequency structure be the key to uniting robust long-horizon planning with precise fine-grained closed-loop control in robotic policies?}

In this work, we present HiPolicy: a hierarchical multi-frequency action chunking framework for policy learning. As shown in~\Fref{fig:different_methods}, our framework jointly predicts action sequences at multiple frequencies from hierarchically aligned observation histories, fusing their representations to model long-horizon and fine-control behavior concurrently. To resolve the execution trade-off between speed and precision, we propose an action-entropy-guided adaptive execution strategy: 
Low-entropy frames indicate concentrated action distributions and stable predictions, prompting execution of high-frequency actions to enable precise closed-loop refinement. 
High-entropy frames reflect dispersed distributions and greater uncertainty, corresponding to broader phase-level decisions, where executing low-frequency actions aligns with high-level intent while increasing operational speed \cite{Guo2025DemoSpeedupAV}.
By dynamically selecting execution frequency based on policy uncertainty, our method speeds up the execution while balancing fine-grained control with long-horizon consistency. 

Our contributions are threefold:  
\begin{itemize}
    \item We introduce HiPolicy, a novel hierarchical multi-frequency action chunking framework for policy learning that addresses the trade-off between long-horizon dependency modeling and high-frequency reactive control.
    \item We propose an action-entropy-guided adaptive execution mechanism that dynamically selects action frequency based on policy uncertainty, enhancing both robustness and execution efficiency.
    \item We validate our approach on extensive simulation benchmarks and real-world manipulation tasks, demonstrating consistent improvements in both performance and efficiency across 2D and 3D generative policies.
\end{itemize}
\section{Related Work}

\subsection{Imitation Learning for Manipulation}
Imitation learning (IL) has become a dominant learning paradigm for robots to acquire manipulation skills from human demonstrations \cite{Chi2023DiffusionPV,Zhao2023LearningFB,Ze2024DP3}. Generative models such as Diffusion Models \cite{ho2020denoising, song2021denoising} and Conditional Variational Autoencoders (CVAE) \cite{kim2021conditional, pu2016variational} are the primary drivers of this progress, enabling accurate and robust action distribution modeling \cite{vahdat2021score, song2020improved, pu2016variational, lipman2022flow}. 

Diffusion Policy \cite{Chi2023DiffusionPV} integrates the diffusion process into generating the action sequence of robots, enabling accurate action prediction. 3D Diffusion Policy \cite{Ze2024DP3} and 3D Diffuser Actor \cite{Ke20243DDA} incorporate point cloud representations into the diffusion policy, enhancing the spatial perception ability for manipulation. 
Our Hierarchical Policy proposes a hierarchical modeling method based on different frequencies, which can be easily integrated into Diffusion Policy \cite{Chi2023DiffusionPV} and 3D Diffusion Policy \cite{Ze2024DP3}, compensating their weakness in modeling non-Markov processes by jointly predicting at hierarchical temporal resolutions. Furthermore, our action entropy-guided execution method chooses the suitable execution frequency adaptively, significantly improves the efficiency of manipulation, and relieves the low-speed problem of diffusion-based policies.

\subsection{Hierarchical Manipulation Policy}

Hierarchical structures in the area of computer vision can effectively perceive and integrate environmental information at different semantic levels, thereby enabling the model to take into account both global information and local details, and improving model performance.

In the robot learning community, recent research has sought to integrate hierarchical structures into imitation learning framework. $\mathrm{H}^3$DP \cite{Lu2025H3DPTD} proposes a visuomotor learning framework that explicitly strengthens the coupling between visual features and action generation by hierarchically conditioning on multi-scale visual representations. Reactive Diffusion Policy \cite{xue2025reactive} introduces a hierarchical slow-fast visual-tactile imitation learning algorithm that uses a slow latent diffusion policy for high-level action chunks and a fast asymmetric tokenizer for low-level control to enable quick responses in contact-rich manipulation tasks.
CARP \cite{Gong2024CARPVP} applies a next-scale autoregressive structure that decouples action generation into multi-scale representation learning via an action autoencoder. Dense Policy \cite{Su2025DensePB} introduces a new bidirectionally expanded learning approach for autoregressive policies, employing an encoder-only architecture to hierarchically unfold the action sequence in a coarse-to-fine manner.

However, the hierarchical structure of these policies only models perceptual input information at different levels, and few work studies have been conducted on imitation learning algorithms with a hierarchical structure for different temporal resolutions. In contrast, our Hierarchical Policy features temporal hierarchical modeling, which hierarchically divides the perceptual input and action sequence simultaneously according to different frequencies and trains the model to predict future action sequences at different time scales.

\section{Method}

{\begin{figure*}[t]    
    \centering
    \includegraphics[width=\textwidth]{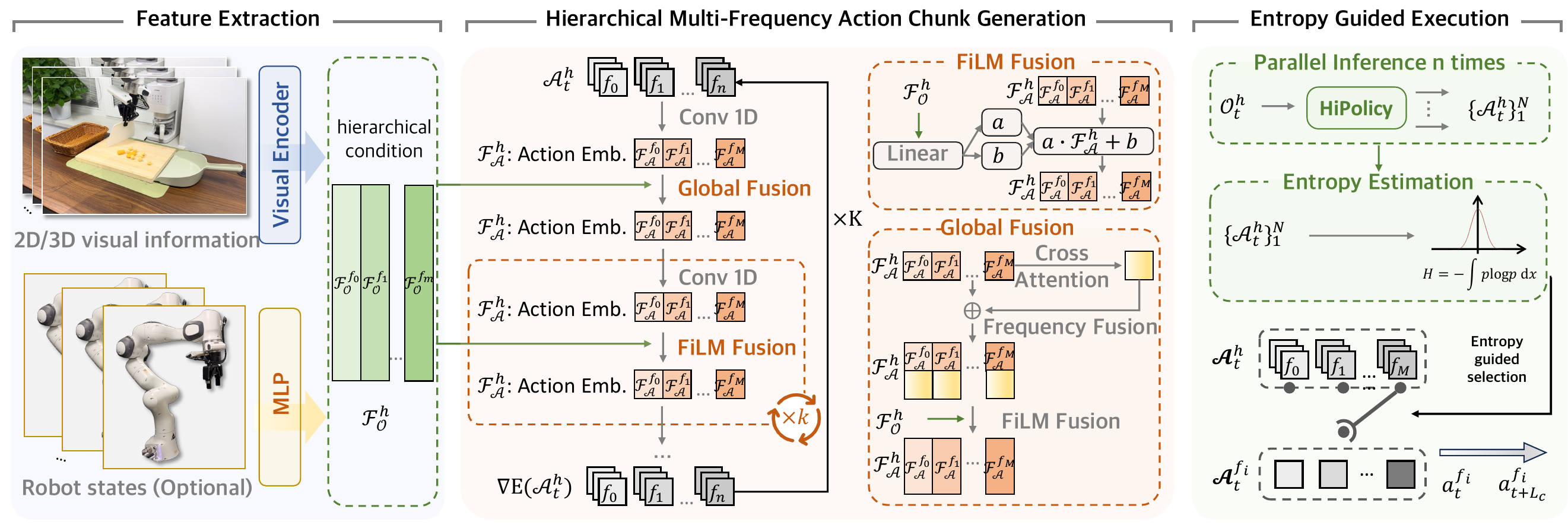}
    \captionof{figure}{
        \textbf{Overview of HiPolicy.} We propose HiPolicy, a hierarchical multi-frequency action chunk policy with an entropy-guided adaptive execution strategy. Given a hierarchical observation history, HiPolicy predicts multi-frequency action chunks simultaneously through a diffusion-based model. During inference, HiPolicy estimates the action entropy through multiple samplings and adaptively chooses the execution frequency according to the estimated entropy.
    }
    \label{fig:overview}
\end{figure*}

\subsection{Problem Formulation}
\label{sec:problem_formulation}

We consider an imitation learning setting where the agent learns a generative policy:
\begin{align}
    \bm{\mathcal{A}}_{t, L_c} = \pi_\theta(\bm{\mathcal{O}}_{t, L_h}),
\end{align}
where $t$ denotes the current time step, and $L_c$ and $L_h$ denote the action chunk length and observation history length, respectively. The policy $\pi$ maximizes the likelihood of the actions given the observation by $P (\bm{\mathcal{A}}_{t, L_c}|\bm{\mathcal{O}}_{t, L_h})$. $\theta$ are the learnable parameters of the policy $\pi_\theta$, and $\bm{\mathcal{O}}_{t, L_h}$ denotes the observation history over $L_h$ time steps:
\begin{equation}
    \bm{\mathcal{O}}_{t, L_h} = \{\bm{o}_{t-L_h+1}, \dots, \bm{o}_t\},
\end{equation}

with $\bm{o}_t = \{v_t, s_t\}$, where $v_t$ denotes the visual observation (either a 2D RGB image or a 3D point cloud) and $\bm{s}_t$ denotes proprioceptive states such as joint positions, velocities, and gripper status. The \emph{action chunk} $\bm{\mathcal{A}}_{t, L_c}$ is defined as:
\begin{equation}
    \bm{\mathcal{A}}_{t, L_c} = \{\bm{a}_t, \bm{a}_{t+1}, \dots, \bm{a}_{t+L_c-1}\},
\end{equation}

where $\bm{a}$ is a low-level control action and $L_c$ is the chunk size, i.e., the number of actions executed before receiving the next observation and generating the subsequent chunk.

\noindent\textbf{Prior formulation (single frequency).}  
In conventional fixed-frequency chunking, both observation histories and action chunks are sampled at a single frequency $f$:
\begin{equation}
    \bm{\mathcal{O}}^f_{t, L_h} = \{\bm{o}^f_{t-L_h+1}, \dots, \bm{o}^f_t\},
\end{equation}

\begin{equation}
    \bm{\mathcal{A}}^f_{t, L_c} = \{\bm{a}^f_t, \bm{a}^f_{t+1}, \dots, \bm{a}^f_{t+L_c-1}\}.
\end{equation}

This formulation poses a challenge in selecting the appropriate frequency $f$: low $f$ supports long-horizon dependency modeling, but reduces opportunities for precise and closed-loop control. High $f$ allows for fine-grained control but limits the field of view for long-horizon tasks.

\noindent\textbf{HiPolicy formulation (hierarchical multi-frequency).}  
HiPolicy jointly encodes multi-frequency observation histories:
\begin{equation}
    \bm{\mathcal{O}}^{\text{h}}_{t, L_h} = 
        \bigcup_{m=1}^M \bm{\mathcal{O}}^{f_m}_{t, L_h},
\end{equation}
and predicts a single hierarchical multi-frequency action chunk:
\begin{equation}
    \bm{\mathcal{A}}^{\text{h}}_{t, L_c} = 
        \bigcup_{m=1}^M \bm{\mathcal{A}}^{f_m}_{t, L_c},
\end{equation}
where each $f_m$ represents a frequency component within the same chunk, and $M$ is the number of frequencies.
Unlike single-frequency approaches, the observation and action components at different frequencies are jointly generated and fused within the chunk, enabling coarse high-level planning and fine-grained reactive control to communicate and synergize.

\noindent\textbf{HiPolicy Overview.} As shown in Figure~\ref{fig:overview}, our HiPolicy is designed as a diffusion-based model, predicting action chunks with hierarchical time resolutions simultaneously. First, we extract hierarchical observation features from visual and proprioception input. Then we obtain the overall action chunk by concatenating noisy action chunks at different frequencies and predict the noise through a 1D-CNN-based U-Net \cite{ronneberger2015u}. Hierarchical FiLM fusion \cite{perez2018film} and global feature fusion are two key fusion modules designed to promote fusing observation features with action and injecting global information across different frequencies. After predicting action chunks at hierarchical temporal resolutions, we perform parallel inference to estimate the action entropy and select the execution frequency according to the entropy adaptively.

We follow Diffusion Policy \cite{Chi2023DiffusionPV} to design our training process, which is detailed described in Appendix~\ref{appendix:training}.

\subsection{Multi-Frequency Action Chunk Prediction}
\noindent\textbf{Hierarchical feature extraction.} To obtain the hierarchical observation feature, we first extract the observation frames of different frequencies from the raw observation chunk. Then the hierarchical feature is obtained through the observation encoder. Input vision observation is selectable between 2D images and 3D point clouds. Robot state is also optional as proprioception information, and is encoded through MLP. 

\noindent\textbf{Hierarchical FiLM fusion.} 
To model observation-action mappings at different frequencies, we use features of different frequencies as conditions for action chunks of different frequencies. As shown in~\Fref{fig:overview}, we adopt the FiLM condition \cite{perez2018film} method and impose the corresponding frequency of observation features to each frequency of action chunk. Then we get the hierarchical action feature $\bm{\mathcal{F}_{\mathcal{A}}}^{\mathrm{h}}$ as follows:
\begin{equation}
    \bm{\mathcal{F}_{\mathcal{A}}}^{\mathrm{h}} \leftarrow \bigoplus_{m=1}^M\mathrm{FiLM}(\bm{\mathcal{F}_\mathcal{O}}^{f_m},\bm{\mathcal{F}_\mathcal{A}}^{f_m})
    \label{film}
\end{equation}

where $\bigoplus$ represents the concatenation on the temporal dimension, $\bm{\mathcal{F}_{\mathcal{A}}}^{\mathrm{h}}\in \mathbb{R}^{B\times (MT) \times C}$ denotes the hierarchical action feature, and $\bm{\mathcal{F}_\mathcal{O}}^{f_m}\in \mathbb{R}^{B \times T \times C}$,$\bm{\mathcal{F}_\mathcal{A}}^{f_m} \in \mathbb{R}^{B \times T \times C}$ are the observation and action feature at frequency $f_m$. $B$ is the batch size, $C$ denotes the hidden feature dimension, and $T$ refers to the temporal dimension. This hierarchical condition design aims to capture feature mapping relations at different temporal resolutions in a corresponding way.

\noindent\textbf{Global feature fusion.} As shown in~\Fref{fig:overview}, we design a global feature fusion to facilitate information communication across different frequencies. 
We first apply frequency fusion, and a cross attention\cite{vaswani2017attention} module is adopted to get the global feature across different frequencies. Cross attention \cite{vaswani2017attention} module is adopted for its performance in capturing context information. Specifically, we add a CLS token \cite{Devlin2019BERTPO} to obtain the global feature token. This process is written as follows.
\begin{equation}
 \bm{\mathcal{F}_\mathcal{A}}^{\mathrm{global}}=\mathrm{CrossAttention}(\bm{\mathcal{F}_\mathcal{A}}^{f_0},\bm{\mathcal{F}_\mathcal{A}}^{f_1},...,\bm{\mathcal{F}_\mathcal{A}}^{f_M})
\end{equation}

where $\bm{\mathcal{F}_\mathcal{A}}^{f_i} \in \mathbb{R}^{B\times T \times C}$ is the local action feature for frequency $f_i$, $\bm{\mathcal{F}_\mathcal{A}}^{\mathrm{global}} \in \mathbb{R}^{B\times T \times C}$ is the global feature across different frequencies. Then we concatenate the global feature and the local feature to obtain hierarchical features.
\begin{equation}
\bm{\mathcal{F}_\mathcal{A}}^{\mathrm{h}}\leftarrow\bm{\mathcal{F}_\mathcal{A}}^{\mathrm{h}}\oplus \bm{\mathcal{F}_\mathcal{A}}^{\mathrm{global}}
\end{equation}

After frequency fusion, we use a FiLM condition, which is performed as \Eref{film}, to promote the fusion of action feature and observation feature.

\subsection{Entropy-Guided Adaptive Execution}

\begin{algorithm}[t]
\DontPrintSemicolon
\caption{Entropy-Guided Execution}
\label{alg:ege}

\SetKwInOut{Input}{Input}
\SetKwInOut{Output}{Output}
\Input{Observation $\bm{\mathcal{O}}^{\text{h}}_{t,L_h}$}
\Output{Action for Execution $\bm{\mathcal{A}}^{\text{exec}}_{t,L_c}$}

$\bm{\mathcal{A}}^{\text{h}}_{t,L_c} \leftarrow \emptyset$\;

\For{$i \leftarrow 1$ \KwTo $N$}{
    $\bm{\mathcal{A}}^{i}_{t,L_c} \leftarrow \pi_{\mathrm{HiPolicy}}(\bm{\mathcal{O}}^{\text{h}}_{t,L_h})$
    \Comment{\textcolor{brown}{Independently sample for N times}}
    $\bm{\mathcal{A}}^{\text{h}}_{t,L_c} \leftarrow \bm{\mathcal{A}}^{\text{h}}_{t,L_c} \cup \bm{\mathcal{A}}^{i}_{t,L_c}$
}

$\bm{\mathcal{A}}^{\text{h}_\text{exec}}_{t,L_c} \leftarrow \bm{\mathcal{A}}^{\text{1}}_{t,L_c}$
\Comment{\textcolor{brown}{Use the first sample output for execution}}

$H \leftarrow
\textit{CalculateEntropy}(\bm{\mathcal{A}}^{\text{h}}_{t,L_c})$
\Comment{\textcolor{brown}{Calculate action entropy}}

$\bm{\mathcal{A}}^{\text{exec}}_{t,L_c} \leftarrow \textit{SelectFrequency}(H,\bm{\mathcal{A}}^{\text{h}_\text{exec}}_{t,L_c})$
\Comment{\textcolor{brown}{Select action according to entropy}}

\Return $\bm{\mathcal{A}}^{\text{exec}}_{t,L_c}$\; 
\end{algorithm}

\begin{wrapfigure}{r}{0.5\textwidth}
    \vspace{-22pt}
    \centering
    \includegraphics[width=0.5\textwidth]{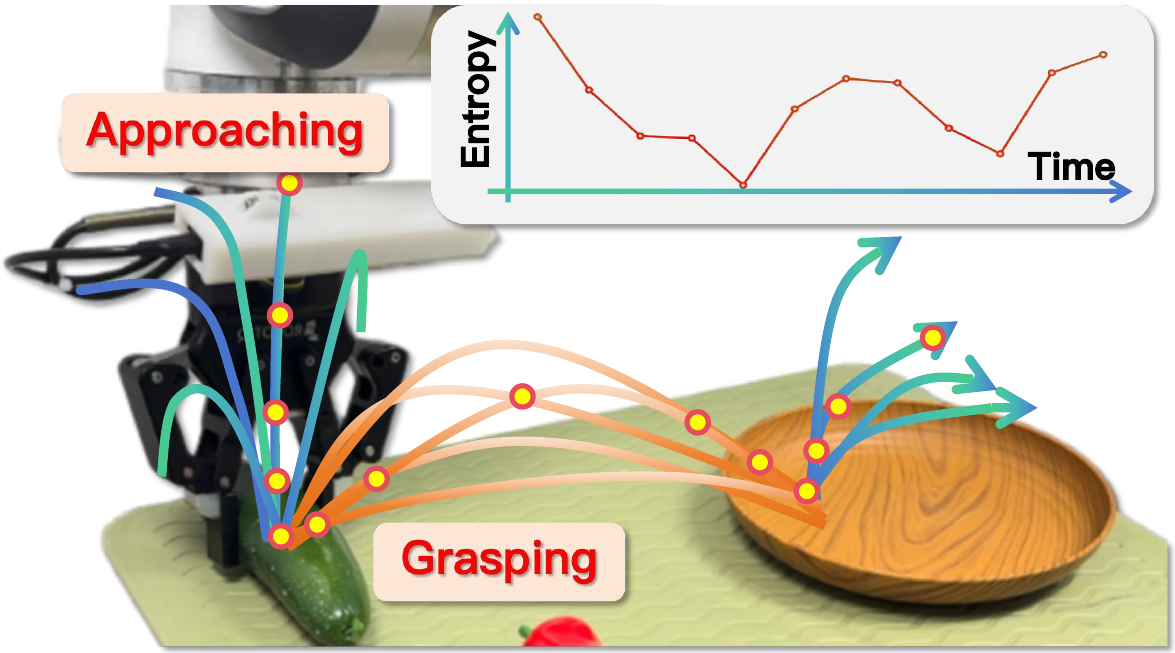} 
    \caption{\textbf{Entropy-Guided Execution.} HiPolicy leverages action entropy, estimated through parallel stochastic inference, as a dynamic gating signal to arbitrate between predictions at multiple frequencies, ensuring a balance between long-horizon planning and reactive control.
    }  
    \vspace{-20pt}
    \label{fig:entropy_exp}
\end{wrapfigure}

To balance the fine-grained closed-loop control and the capability to capture time-dependent information, while speeding up the execution speed of HiPolicy, we propose an entropy-guided execution strategy to perform our policy at an adaptive frequency. As shown in ~\Fref{fig:entropy_exp}, low entropy indicates high consistency (e.g., during grasping), justifying the execution of high-frequency actions for precise, closed-loop control. Conversely, high entropy signals casual motion (e.g., during approaching); here, adaptively switching to low-frequency actions enables high-level planning and significantly accelerates execution. The complete algorithm of entropy-guided execution is shown in Algorithm~\ref{alg:ege}.

To obtain the entropy of an action chunk, we first sample in parallel for $N$ times independently to get $N$ predicted actions when evaluating our HiPolicy, described in the right part of Figure~\ref{fig:overview}. Considering action chunks predicted at a certain frequency, we model the $j$-th action in the chunk as a continuous variable following a Gaussian distribution with probability distribution function:
\begin{equation}
 \hat{p}(\bm{a}^{j}|\bm{o}) = \frac{1}{\sqrt{2\pi}\hat{\sigma}^{j}} \mathrm{exp}(-\frac{(\bm{a}^{j}-\bar{\bm{a}}^{j})^2}{2(\hat{\sigma}^{j})^2})
\end{equation}
where the mean and variance are calculated as follows:
\begin{equation}
\bar{\bm{a}}^{j} = \frac{1}{N}\sum_{i=1}^N\bm{a}_i^{j}
\end{equation}
\begin{equation}
\hat{\sigma}^{j}=\frac{1}{N-1}\sum_{i=1}^N(\bm{a}_i^{j}-\bar{\bm{a}}^{j} )^2
\end{equation}

We use Shannon entropy \cite{shannon1948mathematical} to measure the entropy of the predicted action:
\begin{equation}
\hat{H}^{j} = -\int \hat{p}(\bm{a}^{j}|\bm{o})\mathrm{log}(\hat{p}(\bm{a}^{j}|\bm{o})) \mathrm{d}\bm{a}^{j}
\end{equation}

The simplified formula of action entropy is as follows:
\begin{equation}
\hat{H}^{j} = \mathrm{log}(\sqrt{2\pi\mathrm{e}} \cdot \hat{\sigma}^{j})
\label{equation:entro}
\end{equation}

Then the overall action entropy at time $t$ is the average entropy of all frequencies and all time steps:
\begin{equation}
\hat{H}_t = \frac{1}{ML_c}\sum_{m=1}^M \sum_{j=t}^{t+L_c-1} \hat{H}_t^{f_m,j}
\end{equation}
where $\hat{H}_t^{f_m,j}$ denotes action entropy for $j$-th action in the chunk at frequency $f_m$, calculated following Equation~\ref{equation:entro}.

After estimating the action entropy, we choose execution frequency by comparing the estimated entropy with preset ascending thresholds $\{\hat{H}_{\mathrm{th},1},...,\hat{H}_{\mathrm{th}, M+1}\}$, with frequencies sorted in descending order:
\begin{align}
f_i = f_k, \ \mathrm{if} \ H_t \in (H_{\mathrm{th},k},H_{\mathrm{th},k+1})
\end{align}

where $f_i$ is the selected frequency. Finally, the execution action chunk is:
\begin{align}
\bm{\mathcal{A}}_{t,L_c}^{\text{exec}}=\{\bm{a}_1^{f_i,t}, \dots, \bm{a}_1^{f_i,t+L_c-1}\},
\end{align}

\section{Experiments}
\label{sec:experiments}

\begin{table*}[t]
    \centering
    \caption{\textbf{Detailed evaluation results on RoboTwin 1.0 and RoboTwin 2.0 Benchmark.} Results (success rate) are based on 100 evaluation episodes, distinguishing between precise and other tasks. Bold red highlights the relative percentage improvement over baseline methods, while color shades are used to categorize tasks by their horizon (short, long, and super long). Tasks are sorted according to demo steps. }
    \label{tab:results_robotwin_1_0}
    \adjustbox{max width=1.0\textwidth}{\begin{tabular}{l c  c c | c >{\columncolor{blue!6}} c | c >{\columncolor{blue!6}} c }
    \toprule
    \textbf{Task name}  & \textbf{Task Source} & \textbf{Precise} & \textbf{Demo steps} & \textbf{DP~\cite{Chi2023DiffusionPV}} & \textbf{DP+HiPolicy} & \textbf{DP3~\cite{Ze2024DP3}} & \textbf{DP3+HiPolicy}\\
    \midrule
    \midrule
    \textit{Blocks stack hard}          & RoboTwin 1.0 & \Yes & \textcolor{C1}{665} & 0  & \textbf{10}  & 4  & \textbf{7}   \\
    \textit{Blocks stack easy}          & RoboTwin 1.0 & \Yes & \textcolor{C1}{437} & 0  & \textbf{15}  & 27 & \textbf{44}  \\
    \textit{Block handover}             & RoboTwin 1.0 & \Yes & \textcolor{C1}{415} & 61 & \textbf{100} & 45 & \textbf{99}  \\
    \textit{Dual shoes place}           & RoboTwin 1.0 & \Yes & \textcolor{C2}{384} & 4  & \textbf{15}  & 13 & \textbf{14}  \\
    \textit{Shoe place}                 & RoboTwin 1.0 & \Yes & \textcolor{C2}{251} & 33 & \textbf{48}  & 49 & \textbf{66}   \\
    \textit{Place bread basket}         & RoboTwin 2.0 & \Yes & \textcolor{C2}{231} & 41 & \textbf{64} & 15 & \textbf{47}    \\
    \textit{Block hammer beat}          & RoboTwin 1.0 & \Yes & \textcolor{C2}{211} & 0  & \textbf{67}  & 57 & \textbf{94}  \\
    \textit{Scan object}                & RoboTwin 2.0 & \Yes & \textcolor{C3}{170} & 21 & \textbf{44} & 12 & \textbf{52}   \\
    \textit{Place bread skillet}        & RoboTwin 2.0 & \Yes & \textcolor{C3}{162} & 38 & \textbf{55} & 45 & \textbf{58}   \\
    \textit{Stamp seal}                 & RoboTwin 2.0 & \Yes & \textcolor{C3}{151} & 5 & \textbf{19} & 15 & \textbf{74}    \\
    \textit{Place mouse pad}            & RoboTwin 2.0 & \Yes & \textcolor{C3}{149} & 13 & \textbf{29} & 3 & \textbf{15}    \\
    \textit{Place phone stand}          & RoboTwin 2.0 & \Yes & \textcolor{C3}{130} & 45 & \textbf{79} & 55 & \textbf{64}   \\
    \midrule
    \textbf{Average} & & & & 22 & \textbf{45\space\red{(105\%$\uparrow$)}} & 28 & \textbf{53\space\red{(89\%$\uparrow$)}} \\
    \midrule
    \midrule
    \textit{Put bottles dustbin}    & RoboTwin 2.0 & \No & \textcolor{C1}{637} & 38 & \textbf{81} & 66 & \textbf{80}   \\
    \textit{Stack bowls three}      & RoboTwin 2.0 & \No & \textcolor{C1}{476} & 80 & \textbf{88} & 84 & \textbf{86}   \\
    \textit{Empty cup place}        & RoboTwin 1.0 & \No & \textcolor{C2}{269} & 89 & \textbf{95}  & 54 & \textbf{92}  \\
    \textit{Open laptop}            & RoboTwin 2.0 & \No & \textcolor{C2}{258} & 69 & \textbf{86} & 82 & \textbf{86}   \\
    \textit{Bottle adjust}          & RoboTwin 1.0 & \No & \textcolor{C3}{194} & 46 & \textbf{52}  & 81 & \textbf{85}  \\
    \textit{Dual bottles pick hard} & RoboTwin 1.0 & \No  & \textcolor{C3}{192} & 60 & \textbf{72} & 53 & \textbf{60}  \\
    \textit{Diverse bottles pick}   & RoboTwin 1.0 & \No & \textcolor{C3}{182} & 24 & \textbf{65}  & 44 & \textbf{50}  \\
    \textit{Dual shoes pick easy}   & RoboTwin 1.0 & \No & \textcolor{C3}{179} & 82 & \textbf{94}  & 56 & \textbf{89}  \\
    \textit{Pick apple messy}       & RoboTwin 1.0 & \No & \textcolor{C3}{167} & 40 & \textbf{81}  & \textbf{10} & 3   \\

    \midrule
    \textbf{Average} & & & & 59 & \textbf{79\space\red{(34\%$\uparrow$)}} & 59 & \textbf{70\space\red{(19\%$\uparrow$)}} \\
    \midrule
    \textbf{Total average}                & & & & 37 & \textbf{60\space\red{(62\%$\uparrow$)}} & 41 & \textbf{59\space\red{(44\%$\uparrow$)}} \\
    \bottomrule
\end{tabular}}
\end{table*}

\subsection{Experiments on Simulation Benchmarks}
\label{sec:sim_benchmarks}
\noindent\textbf{Benchmarks and demonstrations.} 
We evaluate the HiPolicy on three widely-used robotic manipulation simulation benchmarks covering a diverse set of tasks ranging from fine-grained to long-horizon manipulation:
\begin{itemize}
    \item \textbf{RoboTwin 1.0} \cite{Mu_2025_CVPR}: RoboTwin 1.0 is a digital twin framework introduced to address the scarcity of diverse demonstration data.  We remove some simple tasks that baseline policies are fully capable of, and evaluate HiPolicy on \textbf{12} challenging tasks from the RoboTwin 1.0 benchmark.
    \item \textbf{RoboTwin 2.0} \cite{chen2025robotwin}: RoboTwin 2.0 is a scalable framework for the automated generation of realistic synthetic data. We removed tasks with fewer than 100 demo steps and with low precision requirements (such as simple pick-and-place) and evaluated HiPolicy on \textbf{9} selected long-horizon or precise tasks from the RoboTwin 2.0 benchmark.
\end{itemize}

To highlight HiPolicy’s ability to jointly model long-horizon dependencies and high-precision closed-loop controls shown in Table\ref{tab:results_robotwin_1_0}, tasks are further categorized, and we present various temporal dependency profiles and levels of control granularity (precise tasks or non-precise tasks). Precise tasks require tight end‑effector pose tolerances to succeed (e.g., cube stacking, QR‑code scanning), while non-precise tasks have more lenient success criteria. 

All demo data was collected according to the methods provided by the official. Details of the selected tasks, demo collection, and preprocessing are provided in Appendix ~\ref{appendix:sim_data}.

\noindent\textbf{Baselines.} 
We compare our method against two representative generative imitation learning policies, selected to cover different input modalities:

\begin{itemize}
    \item \textbf{Diffusion Policy (DP)}~\cite{Chi2023DiffusionPV}: A conditional generative policy that operates purely on 2D image observations, predicting continuous actions via a diffusion process. This baseline reflects the vision-only setting where the policy relies entirely on visual inputs.
    \item \textbf{DP3}~\cite{Ze2024DP3}: Extends diffusion-based policies to 3D perception, capturing the multi-modal setting typical in 3D manipulation, leveraging both color texture cues and geometric shape information.
\end{itemize}

\noindent\textbf{Evaluation protocols.} 
We follow the standard evaluation protocols defined by RoboTwin 1.0 for the tasks from RoboTwin 1.0 \cite{Mu_2025_CVPR} and RoboTwin 2.0 \cite{chen2025robotwin}. Training is conducted for a fixed number of epochs using identical data splits for all methods, ensuring fair comparison. For RoboTwin 1.0 and 2.0, we report \textit{success rate} averaged over $N=100$ episodes per task.

\noindent\textbf{Hyperparameters.} In this paper, we use the unified hyperparameter provided in Appendix ~\ref{appendix:impl_details} for all simulation and real-world tasks.

\subsection{Results and Analysis in Simulation}
\label{sec:results_analysis}

As the simulation results shown in Table~\ref{tab:results_robotwin_1_0}, HiPolicy shows significant relative improvements compared with DP and DP3 baselines. Specifically, in RoboTwin 1.0 and RoboTwin 2.0, it shows a dramatic relative advantage of \textbf{62\%} and \textbf{44\%} versus DP and DP3, respectively.

The most significant improvement appears in tasks such as \textit{block hammer beat} and \textit{stamp seal}, which are representatives of high-precision tasks; \textit{put bottles dustbin}, which is challenging for its super-long horizon properties. These experimental results strongly demonstrate that our HiPolicy successfully reconciles the often-conflicting objectives of maintaining high control accuracy and enabling long-horizon sequential prediction within a single architecture.

Operations such as \textit{blocks stack hard} and \textit{place bread basket} possess both long-horizon and high-precision features. Results on these tasks validate that HiPolicy, whether based on DP or DP3, dramatically enhances the precision of manipulation and improves the modeling capability for long-horizon time-dependent tasks by simultaneously predicting action chunks at hierarchical temporal resolutions.

\subsection{Ablation Study}
\label{sec:ablation}

We conduct ablation studies to isolate the contributions of each component:

\begin{table}[t]
    \centering
    \caption{\textbf{Ablation Experiment.} In the RoboTwin 1.0 platform, we compare our complete HiPolicy with that removing the hierarchical condition, including all conditioned only with the highest frequency (High) or the lowest (Low) frequency, the fusion module, and the hierarchical frequency structure (Hier.), respectively.}
    \label{tab:abl}
    \vspace{-8pt}
    \adjustbox{max width=\linewidth}{\begin{tabular}{l >{\columncolor{blue!6}}c c c c c}
    \toprule
    \multirow{2}{*}{\textbf{Task name}} & \multicolumn{3}{c}{\textbf{Condition}} & \multirow{2}{*}{\textbf{w/o Fusion}}  & \multirow{2}{*}{\textbf{w/o Hier.}} \\
    \cmidrule{2-4}
    & \textbf{Hier.} & \textbf{Low} & \textbf{High} & & \\ 
    \midrule
    \midrule
    \textit{Blocks stack hard} & 10 & 8 & 4 & 8 & 0  \\
    \textit{Blocks stack easy} & 15 & 12 & 10 & 8 & 0 \\
    \textit{Shoe place} & 48 & 43 & 14 & 43 & 33 \\
    \textit{Dual shoes place} & 15 & 12 & 8 & 11 & 4  \\
    \textit{Block handover} & 100 & 100 & 89 & 98 & 61 \\
    \textit{Block hammer beat} & 67 & 66 & 42 & 56 & 0 \\
    \midrule
    \textit{Dual bottles pick hard} & 72 & 70 & 62 & 67 & 60  \\
    \textit{Dual shoes pick easy} & 94 & 94 & 94 & 93 & 82  \\
    \textit{Pick apple messy} & 81 & 80 & 67 & 66 & 40 \\
    \textit{Empty cup place} & 95 & 96 & 95 & 92 & 89 \\
    \textit{Bottle adjust} & 52 & 52 & 50 & 46 & 46\\
    \textit{Diverse bottles pick} & 65 & 61 & 54 & 61 & 24  \\
    \midrule
    \textbf{Average} & \textbf{60} & 58 & 49 & 54 & 37 \\
    \bottomrule
\end{tabular}}
    \vspace{-10pt}
\end{table}

\noindent\textbf{No hierarchical observation condition.} 
The hierarchical condition is replaced by a fixed-frequency condition, where the policy is conditioned solely on high-frequency or low-frequency observations. As shown in Table~\ref{tab:abl}, taking observations at a fixed frequency leads to a performance decline, with more obvious drop when taking a higher fixed frequency of observation.

\noindent\textbf{No hierarchical feature fusion.} 
Remove the attention module for fusing features across different frequencies. From the results in Table~\ref{tab:abl}, the hierarchical feature fusion part promotes feature communication across different frequency chunks, contributing to performance improvement by 6\%.

\begin{wraptable}{r}{0.6\linewidth}
    \centering
    \vspace{-34pt}
    \caption{\textbf{Acceleration Experiment.} We measure the success rate (SR) and the execution steps for DP baseline, DP+HiPolicy without entropy-guided (EG) execution, and DP+HiPolicy across 8 tasks from the RoboTwin 2.0 platform. We use red to highlight our improvement against DP in the success rate and execution speed.}
    \label{tab:acc}
    \adjustbox{max width=\linewidth}{\begin{tabular}{l | C{1.2cm} C{1.2cm} | C{1.6cm} C{1.6cm} | >{\columncolor{blue!6}}C{1.2cm} >{\columncolor{blue!6}}C{1.2cm}}
    \toprule
    \multirow{2}{*}{\textbf{Task name}} & \multicolumn{2}{c|}{\textbf{DP~\cite{Chi2023DiffusionPV}}} & \multicolumn{2}{c|}{\textbf{DP+HiPolicy w/o EG}} & \multicolumn{2}{c}{\textbf{DP+HiPolicy}} \\
    \cmidrule(lr){2-3} \cmidrule(lr){4-5} \cmidrule(lr){6-7}
    & \textbf{SR$\uparrow$} & \textbf{Steps$\downarrow$} & \textbf{SR$\uparrow$} & \textbf{Steps$\downarrow$} & \textbf{SR$\uparrow$} & \textbf{Steps$\downarrow$} \\
    \midrule
    \midrule
    \textit{Stamp seal} & 5 & 140 & 19 & 189 & 25 & 140\\
    \textit{Place mouse pad} & 13 & 164 & 29 & 150 & 28 &90\\
    \textit{Scan object} &  21 & 146 & 44 & 139 & 37 & 103\\
    \textit{Place phone stand} & 45 & 103 & 79 & 102 & 73 & 82\\
    \textit{Place bread skillet} & 38 & 114 & 55 & 113 & 42 & 85\\
    \midrule
    \textbf{Average} & 24 & 133 & 45 & 139 & 
     41\red{(71\%$\uparrow$)} & 100\red{(25\%$\uparrow$)} \\
    \midrule
    \bottomrule
\end{tabular}
}
    \vspace{-25pt}
\end{wraptable}

\noindent\textbf{Single-frequency chunking.}
Remove multi-frequency actions, only predicting and executing actions at a fixed frequency. As the core design of our HiPolicy, removing the hierarchical frequency structure causes significant drop in success rate, as evidenced by the last column of Table~\ref{tab:abl}.

\noindent\textbf{No entropy guidance execution.} 
Entropy-guided execution provides a clear advantage. As shown in Table~\ref{tab:acc}, the strategy improves execution speed by 25\% while incurring only a very small drop in success rate, relative to DP+HiPolicy without entropy-guided execution.

\begin{wraptable}{r}{0.47\linewidth}
    \vspace{-35pt}
    \centering
    \caption{\textbf{Sample times ablation.}} 
    \label{tab:sample_times_ablation}
    \adjustbox{max width=\linewidth}{\begin{tabular}{l c  c | c  c}
    \toprule
    \textbf{\textit{N}} & \textbf{Time(ms)$\downarrow$} & \textbf{VRAM(MB)$\downarrow$} &  \textbf{SR(\%)$\uparrow$} & \textbf{Steps$\downarrow$} \\
    \midrule
    \midrule
    1 & 105.5 & 7394 & -- & -- \\
    5 & 106.0 & 7403 & 35 & 104 \\
    10 & 106.3 & 7414 & 36 & 104  \\
    \cellcolor{blue!10}100 &  \cellcolor{blue!10}107.5 & \cellcolor{blue!10}7457 & \cellcolor{blue!10}41 & \cellcolor{blue!10}100 \\
    500 & 147.9 & 7629 & 39 & 103 \\
    \bottomrule
\end{tabular}}
    \vspace{-22pt}
\end{wraptable}

\noindent\textbf{Sample Times.} 
In this paper, we set $N = 100$ as the unified hyperparameter and inference in parallel. Since the condition features are extracted once and reused, increasing $N$ from 1 to 100 introduces only 2.0 ms of latency and 63 MB of GPU memory overhead, as shown in Table \ref{tab:sample_times_ablation}, which is negligible. Increasing $N$ improves the accuracy of entropy estimation and the task success rate, and saturates around $N=100$. Furthermore, since we use 15Hz as the fixed control frequency for execution, the Wall-Clock time (inference \& execution) is proportional to the action steps.

In general, all hierarchical designs and the entropy-guided execution strategy is essential to balance the long-horizon modeling ability and execution efficiency.

\subsection{Real-world Experiments}
\label{sec:real_robot}

\begin{figure*}[h]
    \vspace{-10pt}
    \centering
    \includegraphics[width=\textwidth]{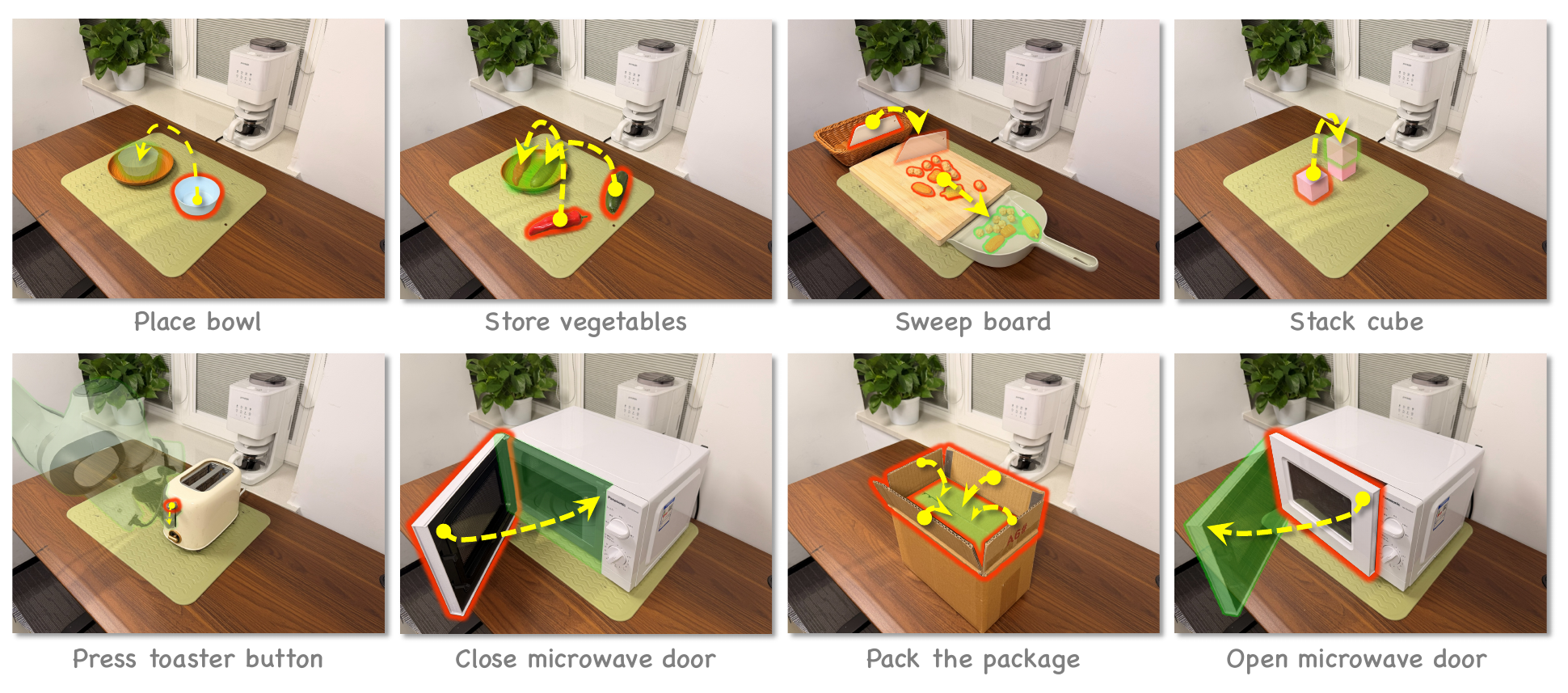}
    \caption{\textbf{Real-world Robot Tasks.} We evaluate our Hierarchical Policy on 8 real-world manipulation tasks with the Franka Panda robot arm.}
    \label{fig:real_robot_tasks}
    \vspace{-10pt}
\end{figure*}

\noindent\textbf{Setup.} 
We use a Franka Emika Panda robotic arm equipped with a Robotiq 2F-85 gripper for our experiments. The setup includes two Zed 2i cameras positioned to provide third-person views and one Zed Mini camera mounted on the robot's end-effector to capture first-person perspectives of the environment, following the setup in DROID \cite{khazatsky2024droid}. Detailed information can be found in Appendix~\ref{appendix:real_robot_setup}.

\noindent\textbf{Tasks.}
As shown in Fig.~\ref{fig:real_robot_tasks}, we evaluate our Hierarchical Policy on eight real-world manipulation tasks:
\begin{itemize}
    \item \textbf{Place Bowl}: The robot is required to grasp the edge of the light blue bowl precisely and place it on top of the plate.
    \item \textbf{Close Microwave Door}: The robot's end effector pushes the door of the white microwave oven, closing it tightly.
    \item \textbf{Store Vegetables}: The robot needs to accurately pick up the chili peppers and cucumbers and place them on a plate. This task serves as an effective test of the policy's viability over a long-horizon manipulation.
    \item \textbf{Pack the Package}: The robot uses its end effector to fold 4 edges of the box, and this task can also effectively test the performance of a policy in long-horizon manipulation.
    \item \textbf{Sweep Board}: The robot is required to grasp a plate and sweep the snacks into the dustpan accurately. The entire process effectively evaluates the robot's performance in fine-grained and long-horizon manipulation tasks.
    \item \textbf{Stack Cube}: The robot needs to place one pink block on top of another pink block stably, which poses a significant challenge to the robot's manipulation precision.
    \item \textbf{Press Toaster Button}: The robot needs to press the very small toaster button all the way down, which needs robot to perform at extremely high manipulation precision.
    \item \textbf{Open Microwave Door}: Initially, the microwave oven door has only a tiny opening, and the robot must precisely open the door within an extremely limited area to succeed.
\end{itemize}

\begin{table}[h]
    \centering
    \caption{\textbf{Detailed evaluation results on 8 real-world tasks.} Our results are reported based on an average of 10 evaluation trials per task, each conducted with randomized object initializations to ensure statistical robustness. And the bold red color is used to highlight the relative percentage improvement over baseline methods. In \textit{close microwave oven}, we respectively report the execution length of closing the door to nearly-closed position and closing the door tightly. } 
    \label{tab:results_transposed}
    \adjustbox{max width=1.0\textwidth}{\begin{tabular}{l| c >{\columncolor{blue!6}} c | c >{\columncolor{blue!6}} c}
    \toprule
    & \multicolumn{2}{c|}{\textbf{Success Rate}$\uparrow$} & \multicolumn{2}{c}{\textbf{Avg. Execution Steps}$\downarrow$} \\
    \midrule
    \multicolumn{1}{c|}{\textbf{Task name}} & \textbf{DP~\cite{Chi2023DiffusionPV}} & \textbf{DP+HiPolicy} & \textbf{DP~\cite{Chi2023DiffusionPV}} & \textbf{DP+HiPolicy} \\
    \midrule
    \midrule
    \textit{Place bowl} & 70\%  &\textbf{90\%} & 98  & \textbf{87} \\
    \textit{Store vegetables} & 60\%  &\textbf{90\%} & 184  & \textbf{159} \\
    \textit{Sweep board} & \textbf{60\%}  &\textbf{60\%} & 202  & \textbf{185} \\
    \textit{Stack cube} & 70\%  &\textbf{80\%} & 137  & \textbf{108} \\
    \textit{Press toaster button} & 70\%  &\textbf{80\%} & 62  & \textbf{58}\\
    \textit{Open microwave door} & 80\%  &\textbf{90\%} & 117  & \textbf{82} \\
    \textit{Close microwave door} & 0\%  &\textbf{100\%} & 60  & \textbf{58/67}\\
    \textit{Pack the package} & 70\%  &\textbf{90\%} & 173  & \textbf{154} \\

    \midrule
    \multicolumn{1}{c|}{\textbf{Avg.}} & 60\%  &\textbf{85\%\space\red{(42\%$\uparrow$)}} & 129  & \textbf{111/113} \\
    \bottomrule
\end{tabular}}
\end{table}

\begin{figure}[h]
    \centering
    \includegraphics[width=\textwidth]{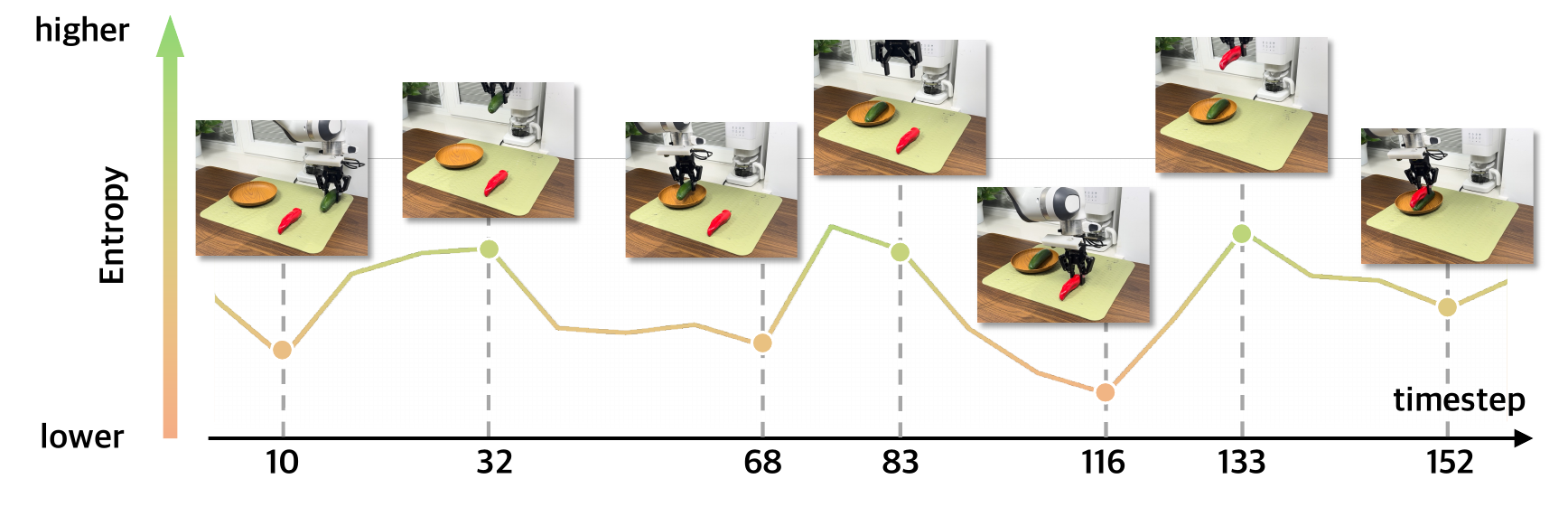}
    \vspace{-12pt}
    \caption{\textbf{Action entropy curve in \textit{store vegetables.} }We display 7 key frames with high or low precision requirements through the task above the entropy curve.}
    \label{fig:entropy}
    \vspace{-12pt}
\end{figure}

\noindent\textbf{Results and analysis in real-world robot experiments.}
As shown in Table~\ref{tab:results_transposed}, HiPolicy shows \textbf{42\%} relatively higher success rate and \textbf{14\%} faster execution speed compared to DP baseline, strongly proving that our HiPolicy balance the execution speed and high control accuracy in challenging real-world tasks. 

The most notable increase is observed in \textit{close microwave door}. The significant difference lies in the fact that closing a microwave oven requires locking its internal latches, while DP only closes the door to a near-closed position without locking the latches, leaving it stuck. Our HiPolicy, however, can perform continuously until the microwave oven is fully closed. Closing the door requires relatively greater force, thus, the demonstration often pauses before closing it tightly, forming a non-Markov process. The experiment result indicates that DP, running at a fixed high frequency (15Hz) to pursue higher control accuracy, fails to keep the long-horizon modeling ability. In contrast, our HiPolicy solves the dilemma between improving manipulation precision and efficiency by modeling with a hierarchical multi-frequency structure.

Tasks such as \textit{store vegetables} and \textit{pack the package}, featuring long-horizon manipulation, also witness significant improvement in the success rate or execution speed of our HiPolicy. These results clearly demonstrate that our HiPolicy successfully balances the execution quality and efficiency in long-horizon tasks by modeling with hierarchical temporal resolutions.

Furthermore, the entropy curve presented in Figure~\ref{fig:entropy} validates our hypothesis that a higher entropy value within a chunk is indicative of actions requiring lower precision (e.g., fetch the vegetable), while a lower entropy value implies high-precision movement (e.g., pick the vegetables).

\section{Conclusion}
We presented HiPolicy, a hierarchical multi-frequency action chunking framework that jointly generates coarse low-frequency and precise high-frequency actions within each chunk, balancing long-horizon dependency modeling and fine-grained closed-loop control. Across diverse simulated and real-world manipulation tasks, HiPolicy consistently improved both performance and efficiency.

\noindent\textbf{Limitations.} Our evaluation is limited to relatively small models and datasets, without exploring integration with large vision-language-action frameworks or other scalable policy architectures. We believe that extending HiPolicy to large-scale multimodal settings could unlock substantial new capabilities for general-purpose embodied agents.


\bibliographystyle{splncs04}
\bibliography{main}
%
%

\clearpage
\clearpage
\setcounter{page}{1}
\setcounter{figure}{0}
\setcounter{table}{0}
\setcounter{section}{0}
\setcounter{equation}{0}
\renewcommand\thesection{\Alph{section}}
\renewcommand\thefigure{\Alph{section}.\arabic{figure}}
\renewcommand\thetable{\Alph{section}.\arabic{table}}

\author{Jiyao Zhang\inst{1, 2} \and
Zimu Han\inst{3} \and
Junhan Wang\inst{1} \and
Xionghao Wu\inst{4} \and
Shihong Lin\inst{5} \and
Jinzhou Li\inst{1} \and
Hongwei Fan\inst{1, 2} \and
Ruihai Wu\inst{1} \and
Dongjiang Li\inst{6} \and
Hao Dong\inst{1, 2}}

\authorrunning{J. Zhang, et al.}
\titlerunning{HiPolicy}

\institute{
\scriptsize
\mbox{CFCS, School of CS, PKU} \and
National Key Laboratory for Multimedia Information Processing, School of CS, PKU \and 
\mbox{XJTU \quad \and THU \quad \and BUAA \quad \and Jingdong Technology Information Technology Co., Ltd} \\
\email{jiyaozhang@stu.pku.edu.cn} \\
\vspace{5pt}
\small
\url{https://hipolicy.github.io}
}

\title{Appendix to HiPolicy: Hierarchical Multi-Frequency Action Chunking for Policy Learning}
\maketitle

\section{Details about Simulation Tasks and Demonstrations}
\label{appendix:sim_data}
\subsection{Demonstration Source.} For both RoboTwin 1.0 and RoboTwin 2.0, we use the official script to collect 100 demonstrations per task automatically.

\subsection{Simulation Tasks.}
We evaluate HiPolicy on 24 simulation manipulation tasks from two benchmarks: RoboTwin 1.0 and RoboTwin 2.0. The tasks are illustrated in Figure \ref{fig:sim_tasks_1.0} and Figure \ref{fig:sim_tasks_2.0}, respectively. Below is a brief description of each task.

\noindent \textbf{Tasks from RoboTwin 1.0:}
\begin{itemize}
    \item \textit{Blocks Stack Hard}: The robot needs to stack 3 cubes that are randomly placed on the table and keep them balanced.
    \item \textit{Blocks Stack Easy}: The requirement for the robot is to stack 2 randomly set-up cubes successfully.
    \item \textit{Block Handover}: The task requires the robot to transfer a miniature cubic object from one gripper to the other.
    \item \textit{Dual Shoes Place}: Utilizing a dual-arm configuration, the robot must simultaneously relocate a pair of shoes to a designated operational zone in front of the robot.
    \item \textit{Shoe Place}: The robot needs to pick up a shoe and place it precisely in a designated area.
    \item \textit{Block Hammer Beat}: The robot is mandated to pick up a hammer and precisely strike a small cube.
    \item \textit{Empty Cup Place}: The robot needs to pick up an empty cup and insert it into another empty cup.
    \item \textit{Bottle Adjust}: The robot must accurately pick up the overturned cup on the table and turn it upright.
    \item \textit{Dual Bottle Pick Hard}: The robot is mandated to grasp two bottles of different types in various positions, and move them to a specific place in front of the robot. 
    \item \textit{Dual Bottle Pick Easy}: The robot needs to simultaneously pick up two bottles of different types, and move them in front of it.
    \item \textit{Diverse Bottles Pick}: The robot needs to pick up two bottles of different types that are laid on the table.
    \item \textit{Pick Apple Messy}: The robot needs to precisely pick up the red apple from the cluttered table.    
\end{itemize}

\begin{figure}[ht]
    \centering
    \includegraphics[width=1.0\textwidth]{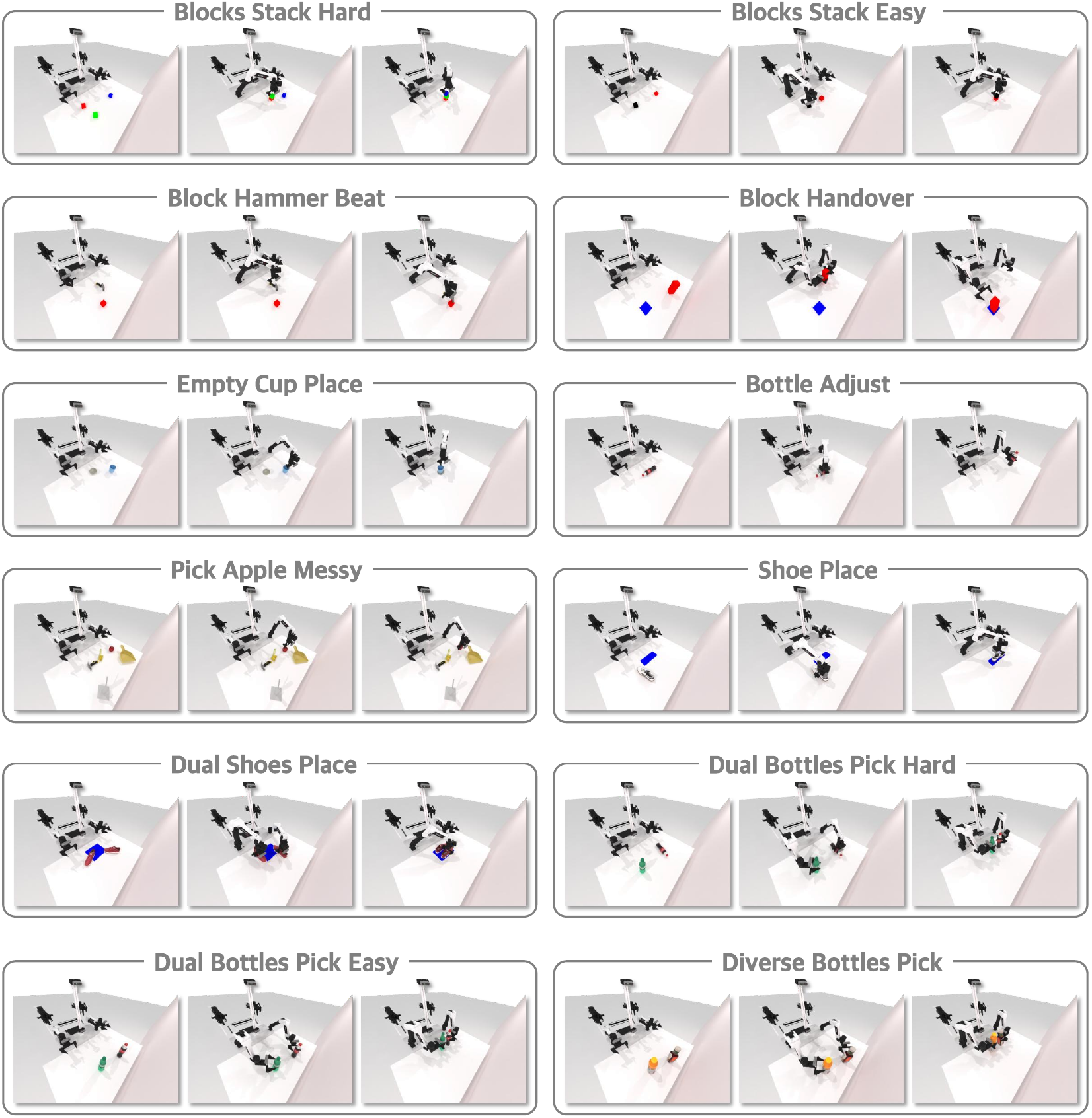}
    \caption{\textbf{Simulation tasks in RoboTwin 1.0 benchmark.} We evaluate HiPolicy on 12 manipulation tasks from the RoboTwin 1.0 benchmark.}
    \label{fig:sim_tasks_1.0}
\end{figure}

\begin{figure}[ht]
    \centering
    \includegraphics[width=1.0\textwidth]{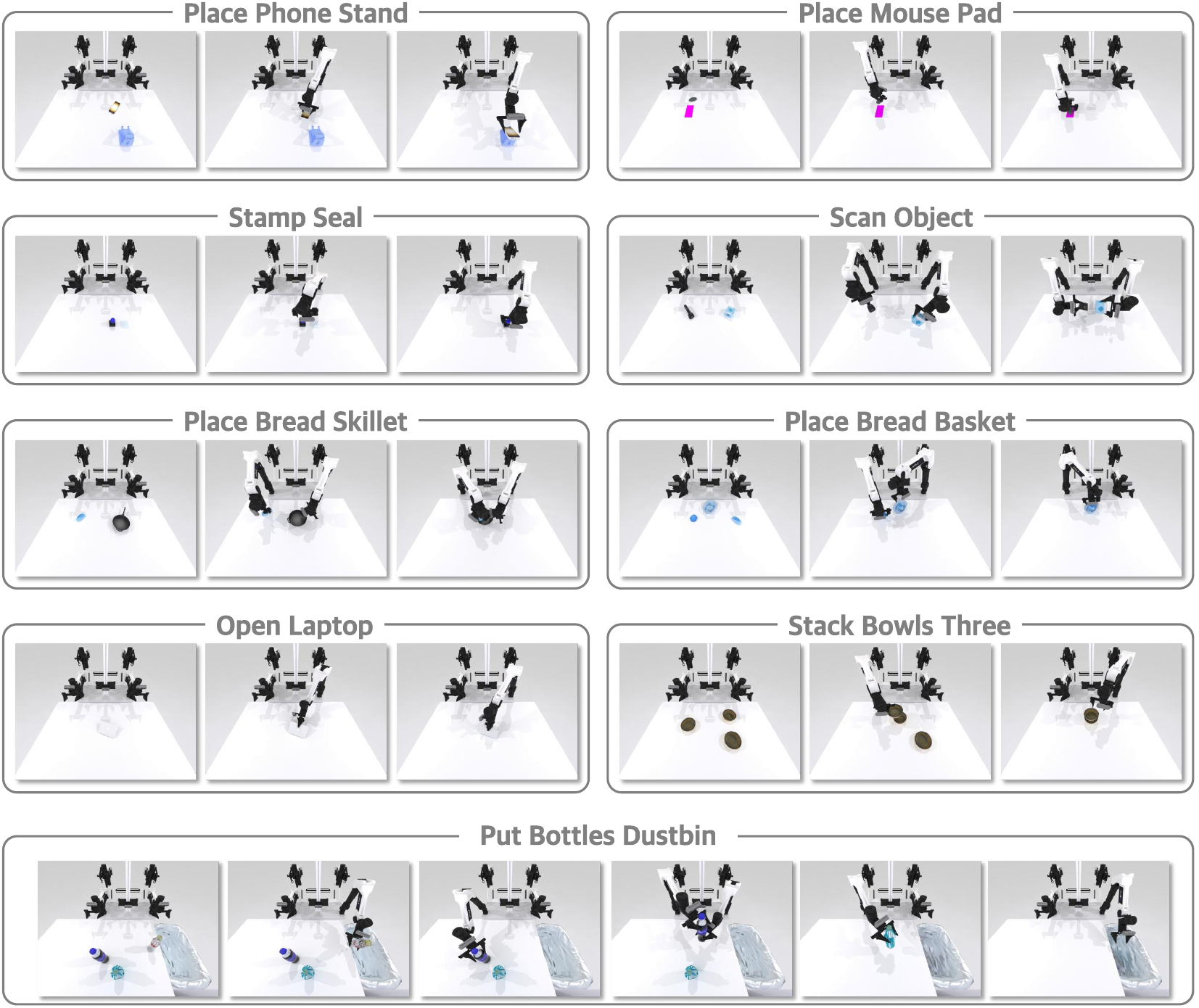}
    \caption{\textbf{Simulation tasks in RoboTwin 2.0 benchmark.} We evaluate HiPolicy on 9 manipulation tasks from the RoboTwin 2.0 benchmark.}
    \label{fig:sim_tasks_2.0}
\end{figure}

\noindent \textbf{Tasks from RoboTwin 2.0:}
\begin{itemize}
    \item \textit{Place Bread Basket}: The robot needs to use both hands to grab two pieces of bread and put them into the basket.
    \item \textit{Scan Object}: The robot needs to use one hand to grasp an object on the table and the other hand to grasp a scanner, then point the scanner at the QR code on the object.
    \item \textit{Pack Bread Skillet}: The robot is mandated to hold the skillet with one hand and a piece of bread with the other, then place the bread into the skillet.
    \item \textit{Stamp Seal}: The robot needs to pick up the tiny stamp on the table and stamp it on a piece of paper.
    \item \textit{Place Mouse Pad}: The robot needs to grab the mouse placed on the desktop and put it in the designated area.
    \item \textit{Place Phone Stand}: The robot is required to grab the phone on the table and place it on the stand.
    \item \textit{Put Bottles Dustbin}: The robot is required to put three blankets that are scattered on the table into a storage bag next to the table.
    \item \textit{Stack Bowls Three}: The robot must stack three bowls that are scattered on the table, keeping them balanced.
    \item \textit{Open Laptop}: The robot needs to open a closed laptop on the table.
\end{itemize}

\section{Detailed Evaluation Protocols}
\label{appendix:eval_prot}
Our evaluation settings for each benchmark are based on the protocols proposed by the respective benchmarks. For the RoboTwin benchmark (versions 1.0 and 2.0), we evaluate DP and DP+HiPolicy at epoch 600 over 100 independent episodes, and report the average success rate. For DP3 and DP3+HiPolicy, we follow the DP3 benchmark’s protocol by evaluating at epoch 3000 over 100 episodes, also reporting the average success rate.

\section{HiPolicy Details}
\label{appendix:training}

The generative policy in HiPolicy adopts a conditional denoising diffusion formulation, in which action trajectories are modeled as a sequence of progressively denoised states conditioned on sensory observations $\mathbf{O}$. 
Beginning from an initial Gaussian sample $\mathbf{a}^{K}$, the reverse process iteratively refines the action state through $K$ steps:
\begin{equation}
\mathbf{a}^{k-1} 
= \alpha_{k} \left( \mathbf{a}^{k} - \gamma_{k} \, f_{\theta}(\mathbf{a}^{k}, k, \mathbf{O}) \right) 
+ \sigma_{k} \, \boldsymbol{\epsilon}^{k}, 
\quad \boldsymbol{\epsilon}^{k} \sim \mathcal{N}(\mathbf{0}, \mathbf{I}),
\label{eq:hipolicy_update}
\end{equation}
where $\alpha_{k}$, $\gamma_{k}$, and $\sigma_{k}$ are step-dependent coefficients determined by a predefined diffusion scheduler, and $f_{\theta}$ predicts the noise component at step $k$ conditioned on $\mathbf{O}$.

\medskip
\noindent
\textbf{Forward Noising Process.}
During training, a clean action $\mathbf{a}^{0}$ is perturbed to construct the noisy state $\tilde{\mathbf{a}}^{k}$:
\begin{equation}
\tilde{\mathbf{a}}^{k} 
= \sqrt{\bar{\alpha}_{k}} \, \mathbf{a}^{0} 
+ \sqrt{1 - \bar{\alpha}_{k}} \, \boldsymbol{\epsilon}^{k},
\label{eq:hipolicy_noising}
\end{equation}
where $\bar{\alpha}_{k} = \prod_{i=1}^{k} \alpha_{i}$ represents the cumulative product of scheduler coefficients.

\medskip
\noindent
\textbf{Training Objective.}
The network parameters $\theta$ are optimized to minimize the discrepancy between the injected noise and the network's prediction for $\tilde{\mathbf{a}}^{k}$:
\begin{equation}
\mathcal{L}(\theta) 
= \mathbb{E}_{k \sim \mathcal{U}(1,K),\, \mathbf{a}^{0},\, \boldsymbol{\epsilon}^{k}}
\left[ \left\| \boldsymbol{\epsilon}^{k} - f_{\theta}(\tilde{\mathbf{a}}^{k}, k, \mathbf{O}) \right\|_{2}^{2} \right],
\label{eq:hipolicy_loss}
\end{equation}

\section{Implementation Details}
\label{appendix:impl_details}

\noindent \textbf{Hyperparameters.} The hyperparameters used to train our model are provided in Table ~\ref{tab:hyper}. Exceptionally, the image resolution is $120\times120$ for \textit{Tool Hang} in Robomimic, since this task requires super high precision.

\begin{table}[h]
    \centering
    \caption{\textbf{Hyperparameters in simulation benchmark.}}
    \label{tab:hyper}
    \adjustbox{max width=1.0\textwidth}{\begin{tabular}{l|c}
    \toprule
    \textbf{Hyperparameter} & \textbf{Value} \\
    \midrule
        Batch Size & 128 \\
        Observation Horizon ($L_h$) & 3 \\
        Action Horizon  & 8 \\
        Prediction Action Horizon ($L_c$) & 8 \\
        Optimizer & AdamW \\
        Betas ($\beta_1$, $\beta_2$) & $[0.9, 0.999]$ \\
        Learning Rate & $1.0\mathrm{e}{-4}$ \\
        Weight Decay & $1.0\mathrm{e}{-6}$ \\ 
        Diffusion Step Embedding Dimension & 128 \\
        Inference Step & 100 \\
        Prediction Type & $\epsilon$-prediction \\
        Image Resolution & $320 \times 240$ \\
        Input View & Head Camera \\
        Frequency Number ($M$) & 3 \\
        Sample Times ($N$) & 100 \\
        \bottomrule
\end{tabular}}
\end{table}

\section{More Ablation Experiments}

\noindent\textbf{Threshold Ablation.}
We use thresholds $H_{\mathrm{th}} = \{-\infty, -6.0, -5.5, +\infty\}$ across all tasks, statistically derived from the 10th and 70th entropy percentiles of the \textit{`Scan object'} task, which requires placement error $\le$ 2.5 cm. As shown in Table \ref{tab:threshold_ablation}, higher thresholds slightly improve SR but greatly lower speed, while lower ones speed execution at the cost of SR, confirming our balanced choice.

\begin{table}[h]
    \centering
    \caption{\textbf{Entropy thresholds ablation.}} 
    \label{tab:threshold_ablation}
    \adjustbox{max width=\linewidth}{\begin{tabular}{c | c c c}
    \toprule
    \textbf{\textit{$H_{\mathrm{th}}$}} & \textbf{SR(\%)} & \textbf{Steps} &\textbf{Explanation} \\
    \midrule
    \midrule
    
    \cellcolor{blue!10}$\{ -\infty, -6.0, -5.5, +\infty \}$ & \cellcolor{blue!10}45 & \cellcolor{blue!10}108 & \cellcolor{blue!10}Thresholds used in manuscript\\
    $\{ -\infty, -5.0, -4.5, +\infty \}$ & 47 & 143 & Increase the probability of executing high-freq. actions\\
    $\{ -\infty, -6.5, -6.0, +\infty \}$ & 37 & 75 & Increase the probability of executing low-freq. actions\\
    \bottomrule
\end{tabular}
}
\end{table}

\noindent\textbf{Hyperparameter Ablation.}
As shown in~\Tref{tab:typer_abl}, we conducted additional ablation studies, varying the observation history ($L_h$), chunk size ($L_c$), and frequency levels ($M$) around our default settings ($L_h=3, L_c=8, M=3$). 
Moderate observation and chunk lengths yield better performance, with HiPolicy outperforming the baseline(24\%) in most settings.

\begin{table}[h]
    \centering
    \caption{\textbf{Hyperparameter ablation.}} 
    \label{tab:typer_abl}
    \adjustbox{max width=\linewidth}{\begin{tabular}{l c  c | c}
    \toprule
    \textbf{$L_h$} & \textbf{$L_c$} & \textbf{$M$} &  \textbf{Success Rate(\%)} \\
    \midrule
    \midrule
    \cellcolor{blue!10}3 & \cellcolor{blue!10}8 & \cellcolor{blue!10}3 & \cellcolor{blue!10}47 \\
    1 & 8 & 3 & 30 \\
    5 & 8 & 3 & 41 \\
    3 & 4 & 3 & 41 \\
    3 & 8 & 1 & 24 \\
    \bottomrule
\end{tabular}}
\end{table}

\section{Real-world Robot Experimental Setup}
\label{appendix:real_robot_setup}
Our real-world experiments follow the hardware configuration described in the DROID benchmark. We employ a Franka Emika Panda 7-DoF manipulator with a Robotiq 2F-85 parallel gripper. Visual observations are captured by three cameras from the ZED family: 

\begin{itemize}
    \item Two ZED 2i cameras are placed at fixed third-person viewpoints on the left and right sides of the workspace, providing complementary global scene coverage of object positions, environment context, and overall task progress.
    \item One ZED Mini camera is mounted on the robot’s end-effector, offering an egocentric perspective aligned with the gripper pose to capture fine-grained local interactions between the gripper and manipulated objects.
\end{itemize}

\begin{figure}[ht]
    \centering
    \includegraphics[width=0.48\textwidth]{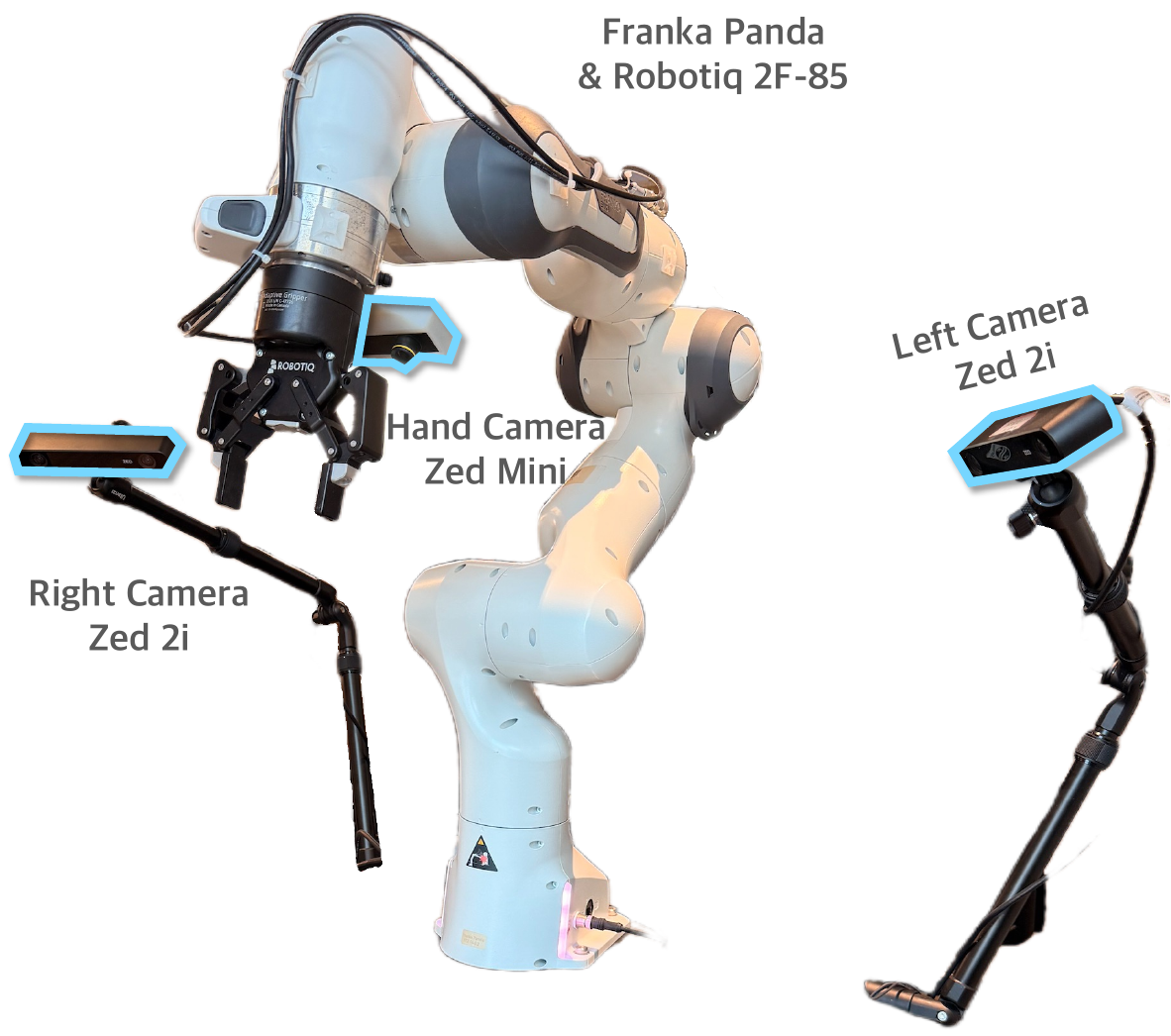}
    \caption{\textbf{Real-world robot experimental setup.} We use a Franka Emika Panda robot equipped with a Robotiq 2F-85 gripper. Two third-person view Zed 2i cameras are used to capture the scene from different angles. A Zed Mini camera is mounted on the robot end-effector to provide an egocentric view. We just use the RGB images from these cameras as the image observations.}
    \label{fig:real_robot_setup}
\end{figure}

This multi-view configuration replicates the observation space design in DROID. All real-world demonstrations in our experiments are collected using the DROID teleoperation system, which enables efficient recording of high-quality trajectories consistent with the benchmark protocol.  

\end{document}